\documentclass[letterpaper]{article} 
\usepackage{aaai24}  
\usepackage{times}  
\usepackage{helvet}  
\usepackage{courier}  
\usepackage[hyphens]{url}  
\usepackage{graphicx} 
\usepackage{natbib}  
\usepackage{caption} 
\usepackage{algorithm}
\usepackage{algorithmic}

\usepackage{newfloat}
\usepackage{listings}
\DeclareCaptionStyle{ruled}{labelfont=normalfont,labelsep=colon,strut=off} 
\floatstyle{ruled}
\newfloat{listing}{tb}{lst}{}
\floatname{listing}{Listing}
\usepackage{amsmath}
\usepackage{amssymb}
\usepackage{booktabs}
\usepackage{textcomp}
\usepackage{gensymb}
\graphicspath{{images/}}
\usepackage{multirow}
\usepackage{arydshln}
\usepackage{color}

\usepackage{pdfpages}

\title{Sequential Modeling of Complex Marine Navigation: Case Study on a Passenger Vessel (Student Abstract)}

\author{
    Yimeng Fan\textsuperscript{\rm 1\equalcontrib}, 
    Pedram Agand\textsuperscript{\rm 1\equalcontrib}, 
    Mo Chen\textsuperscript{\rm 1}, 
    Edward J. Park \textsuperscript{\rm 1}, 
    Allison Kennedy\textsuperscript{\rm 2}, 
    Chanwoo Bae\textsuperscript{\rm 3}
}
\affiliations{
    \textsuperscript{\rm 1}Simon Fraser University, BC, Canada.\\
     \textsuperscript{\rm 2}National Research Council Canada, Ontario, Canada.\\
     \textsuperscript{\rm 3}BC Ferry Inc., BC, Canada.\\
      \{yfa56,pagand, ed\_park\}@sfu.ca, mochen@cs.sfu.ca, allison.kennedy@nrc-cnrc.gc.ca, chanwoo.bae@bcferries.com
}

\usepackage{bibentry}

\begin{document}

\maketitle

\begin{abstract}
The maritime industry's continuous commitment to sustainability has led to a dedicated exploration of methods to reduce vessel fuel consumption. This paper undertakes this challenge through a machine learning approach, leveraging a real-world dataset spanning two years of a ferry in west coast Canada. Our  focus centers on the creation of a time series forecasting model given the dynamic and static states, actions, and disturbances. This model is designed to predict dynamic states based on the actions provided, subsequently serving as an evaluative tool to assess the proficiency of the ferry's operation under the captain's guidance. Additionally, it lays the foundation for future optimization algorithms, providing valuable feedback on decision-making processes.
To facilitate future studies, our code is available at \url{https://github.com/pagand/model_optimze_vessel/tree/AAAI}.
\end{abstract}

\section{Introduction}
\label{sec:intr}
Numerous researchers explore models for vessel Fuel Consumption (FC) prediction using log-based and sensor-based data. Log-based methods can be associate with human errors and suffer from lower sample frequency. 
Sensor-based models, follow a process, including data normalization and feature engineering. 
Nevertheless, a gap exists in the literature as most models lack real-time ship operator input and environmental considerations. Further investigation is needed for feature selection, multicollinearity, non-stationary data handling, and better integration of domain knowledge with data-driven approaches. \citet{agand2023fuel} propose a comprehensive approach integrating domain knowledge and data-driven techniques, incorporating physical insights, correlation matrices, and PCA. However, it does not account for data temporality, which may impact performance.

\begin{figure}
    \centering
    \includegraphics[width=.95\linewidth,trim={0cm  7cm 12cm 0cm},clip]{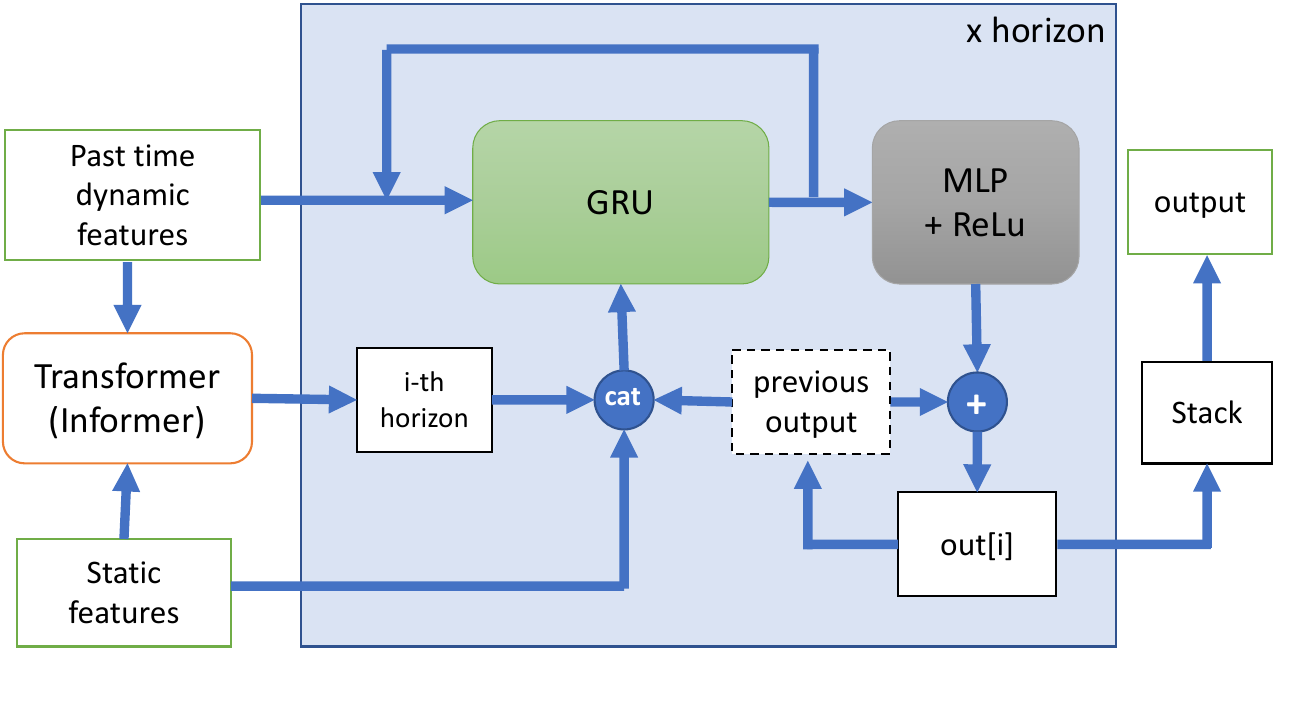}
    \caption{Proposed sequential model architecture}
    \label{fig:architecture}
\end{figure}

In this work, a reality-based time series forecasting model with the primary objective of auto-regressively predicting states of a ferry is proposed. For this analysis, an operational dataset measured on the ferry  every minute over a two year period was used. These records capture the sensor data from a ferry operating in the west coast of Canada for more than 3000 transits along a constant route. This predictive model's pivotal role lies in its ability to generate a precise representation of the ship's states such as location -latitude (LAT) and longitude (LON)-, fuel consumption (FC), and speed over ground (SOG) in future steps. According to Fig. \ref{fig:architecture}, we utilized the previous predicted values in an auto-regressive manner, with information about external disturbance (e.g. wind, current, weather, etc.), and static features (such as direction of movement, docking area, elapsed time of trip, etc). Finally, we open-sourced a reinforcement learning (RL) compatible dataset to D4RL framework  in addition to a Gym environment \cite{fu2020d4rl}. 
\section{Preprocessing}
\label{sec:pre}
In this stage, outlier management was addressed using the 1.5 IQR method, treating them as missing data points and employing various imputation techniques for isolated missing values within trips. 
After devising domain knowledge on draft, cargo, and waves, we developed a clustering method to assign operating modes (mode 1 for autopilot travel and mode 2 for docking regions) based on factors like speed, acceleration, and shaft speed. Wind factors were incorporated through squared relative wind speed and categorized wind direction. Feature selection involved computing the Pearson correlation coefficient and leveraging domain knowledge for decision-making. Feature engineering introduced acceleration and displacement, and we normalized data values to a 0-1 range. Additionally, power transformations were applied to features with skewed distributions, aiming to align them more closely with Gaussian distributions.

\section{Modeling}
\label{sec:mod}
As shown in Fig. \ref{fig:architecture}, the  framework consists of two components: a pretrained transformer model called informer \cite{zhou2021informer}, responsible for executing time series forecasting and feature fusion. Inspired from \cite{agand2023letfuser}, it is followed by a Gated Recurrent Unit (GRU) module employed to predict the residual, thereby mitigating auto-regressive cumulative errors within the predicted outcomes. The determination of the sequence length necessitates two crucial considerations. Tt must be of sufficient to facilitate accurate forecasting while it should avoid excessive elongation, as it signifies the time interval to wait in each trip before making new predictions. Following trade-offs, a sequence length of 25 minutes, coupled with a prediction horizon of 5 minutes, was deemed the optimal configuration. 

\renewcommand{\arraystretch}{0.8}
\begin{table}
\begin{tabular}{ccp{1cm}p{1cm}p{1cm}p{1cm}}
\toprule
&Metric &NAR & NAR +GRU & AR&AR +GRU\\ 
\midrule
 \multirow{3}{*}{FC}&RMSE&0.1876&0.0769&0.0777&\bf{0.0608}\\
  &Std.&0.0856&0.04049& 0.0605&\bf{0.0365}\\
  &$R^2$&-0.023&0.296&0.298&\bf{0.701}\\
  \midrule
 \multirow{3}{*}{SOG}&RMSE&0.1412&0.0322&0.0435&\bf{0.0304}\\
 & Std.&0.0854&\bf{0.0178}&0.0451&0.0179\\
  & $R^2$&0.091&0.602&0.404&\bf{0.793}\\
  \midrule
  \multirow{3}{*}{LAT.}&  RMSE&0.1023&0.0684&0.0613&\bf{0.0385}\\
 & Std.&0.0712&0.0167&\bf{0.0133}& 0.0137\\
  & $R^2$&0.281&0.6193&0.685&\bf{0.874}\\
  \midrule
  \multirow{3}{*}{LON.}& RMSE&0.1370&0.1097&0.0960&\bf{0.0604}\\
 &Std.&0.0842&0.0267&0.0225& \bf{0.0157}\\
  & $R^2$& 0.404&0.671&0.746&\bf{0.899}\\
\bottomrule
\end{tabular}
\caption{Prediction results for non-auto-regressive (NAR) and auto-regressive (AR) w/o GRU refinement.}
\label{tab:cmp}
\end{table}

\section{Environment}
\label{sec:rld}
We have open sourced a RL compatible dataset for offline RL setting with a gym environment that can be served as reality-based simulator. We constructed an offline dataset tailored for conventional RL frameworks, comprising current/next observations, actions, rewards, and termination indicators. The actions encompass heading, shaft speed, and mode selections. For observations, we employ both current and next states, encompassing static factors such as trip start hour, direction, dynamic factors including heading rate, resistance (torque/thrust) \cite{2019577}, displacement, and previous outputs, as well as disturbance-related variables like time, weekday, current, season, weather, wind direction, wind force, and water depth. In terms of rewards, we have defined three distinct intermediate rewards. The first penalizes deviations from the top 1\% trips of the dataset that uses the least fuels, the second penalizes fuel consumption (FC), and the last comprises sparse rewards, with +1 awarded for on-time arrival at the docking area and a penalty of -0.1 for each minute beyond the schedule. The model outputs four variables: LAT, LON, SOG, and FC.

In the Gym environment, we implemented curriculum learning by defining three reward stages. The first stage focuses on emulating the decisions made by captains in the dataset. In the second stage, rewards are structured to encourage behavior similar to the top 1\% of best-performing trips. The third stage emphasizes the minimization of FC while adhering to the designated time schedule. The reinforcement learning process concludes with the issuance of the  ``done'' signal, triggered when the ferry successfully reaches its intended destination. Conversely, if the ferry fails to reach its destination within 25\% more time steps than usual, the process times out, resulting in a negative reward.

\section{Conclusion and Future Work}
\label{sec:futr}
As demonstrated in Table \ref{tab:cmp}, we observe that the root mean square error (RMSE) and coefficient of determination ($R^2$) exhibit the best performance for all four quantities when employing the AR + GRU approach. The second-best results are associated with the standard deviation (Std.) of SOG and LAT, while FC and LON still exhibit the best performance across all approaches. In general, the AR approach outperforms the others, primarily due to allowing the transformer to learn and mitigate cumulative errors during training. 

We outline a step-wise approach to develop a time-series model for a ferry using real data. Additionally, we introduce an offline dataset and a simulator that  both can serve as  training tools for captains or an environment for machine learning systems. The forecasting model will be instrumental in assessing the optimization model's effectiveness. The Gym environment undergoes three stages of training. Initially, it aims to replicate captain driving behaviors, then shifts to imitating top trips with the lowest FC, and finally strives to surpass captain performance by pursuing  to minimize FC while adhering to the time schedule and other limitation. This multi-stage training approach is expected to accelerate the convergence of RL systems. For future direction, we consider  leveraging RL to optimize the navigational best practice to maximize fuel efficiency.


\bibliography{aaai24}
\newpage
\appendix


\section*{Supplementary material}

# Supplementary materials

We will commence with an extended literature review, followed by a detailed exploration of the preprocessing stages. Subsequently, we will delve into the modeling process, including the engineered features. Additionally, we will introduce the publicly available RL dataset, along with a comprehensive explanation of the intermediate reward calculation. Finally, we will introduce the Gym environment and provide a visual comparison between a rendered random trip and the offline dataset. We will render the simulator in both online and auto-regressive modes for a comprehensive analysis which shows how the model can act as a filter.

The International Maritime Organization (IMO) announced their new emissions reduction targets, demanding an approximate 15% reduction in emissions from 2022 to 2030 (IEA 2023). Indeed, transitioning to cleaner energy sources in the maritime industry can be costly and time-consuming. However, an alternative approach that is gaining prominence involves a concerted effort to reduce fuel consumption (FC) (Zhou et al. 2022). This dual objective not only contributes to significant reductions in emissions but also serves as an effective cost-cutting measure for ship operations. However, to achieve this, a reliable model for the vessel is required which is usually limited by accuracy.

# Preprocessing

## Domain knowledge

We have organized the original raw data into trips, corresponding to the vessel being docked at the Bay, as our primary focus lies in data during active travel. Our study looks at significant factors affecting fuel, like ship speed, RPM, draft, trim, cargo, wind, and sea effects (Bal Beşikçi et al. 2016). However, it's worth noting that the dataset does not provide direct information on draft, cargo, and sea effects. To address this, we derive an approach to estimates these missing factors using domain knowledge.

The vessel has two operating modes: Mode 1 for autopilot travel with one engine in operation, and Mode 2 for docking regions with both engines active and controlled by captain. However, the sensor data that directly indicate the mode is absent. Therefore, we have developed a classification approach for mode assignment based on factors such as speed (under or above 16 knots), acceleration (greater or less than 1 knot per square hour), and engine speed (under or above 800 RPM). To encompass the wind force, we square the relative wind speed, and for wind direction, we classify it into forehead, side, and tailwind using relative angles, as recommended in (Panapakidis, Sourtzi, and Dagoumas 2020).

## Feature selection and engineering

The Pearson correlation coefficient for each pair of the two features is first computed. In instances where a high correlation coefficient is observed, we rely on domain knowledge to guide the decision of which feature to remove. As for feature engineering, we introduce acceleration, which signifies alterations in SOG. Moreover, we incorporate displacement, encompassing changes in GPS location, distance traveled, and changes in heading or turns. We conduct data normalization to scales data values in range of 0 to 1. For features exhibiting skewed distributions – like SOG, FC, and engine speed – we additionally apply power transformations. These transformations are aimed at aligning their distributions more closely with Gaussian distributions.

## Outlier detection and handling missing data

We managed outliers using the 1.5 IQR method (Vinutha, Poornima, and Sagar 2018). When data points are removed in this process, we treated them as missing values. To address missing data, we excluded entire trips if consecutive records had excessive missing values and imputed isolated missing values within trips. Imputation methods included using the population mean, trip mean, or interpolating with adjacent records based on the feature's property.

The imputation approach for each field is as follows: given that the vessel's operational area is well-defined, water depth is imputed with the population mode. Heading, wind force, and wind direction are imputed with the mean value within each corresponding trip, as these particular variables exhibit a notable degree of stability within individual trips. Other dynamic and continuous variables are imputed according to the non-missing previous and subsequent values. Imputation was performed through linear interpolation between these adjacent data points. For example, if there is a missing speed, the previous sample value for speed is 800, and the subsequent sample value is 900, we will use 850 to fill in the missing value.

# Modeling

Our model relies on various inputs to make precise predictions, effectively representing the environmental state. These inputs encompass actions - heading (deg), engine speed (rpm), mode (binary) - static state - binary and categorical indicators, including whether the day of the trip falls on a weekday, the direction of the trip, the season in which the trip occurs, and the hour of the trip direction (binary) - observations - turn (deg/min), acceleration (knots/min), water resistance, latitude (LAT) and longitude (LON), both in degrees, SOG (knots), velocity (knots), distance to the destination (degrees), FC (kg/h) - and disturbance factors - time (min), is weekday (categorical), current (knots), season (categorical), hour (categorical), rain (mm), snowfall (mm), wind direction (categorical), and wind force ($knots^2$). The water resistance ratio derived using the following relation (Carlton 2019):

$$\text{Water resistance} = \frac{\text{Thrust}}{\text{Torque} * \text{Speed}} \quad (1)$$

In addition, the velocity vector was computed as follow:

$$\text{Velocity} = [\text{sog} \times \sin(\text{heading}), \text{sog} \times \cos(\text{heading})]^T \quad (2)$$

For categorical features such as season and wind direction, we applied one-hot encoding to represent them numerically. Each unique category is converted into a binary vector, where each category is assigned a separate binary column.

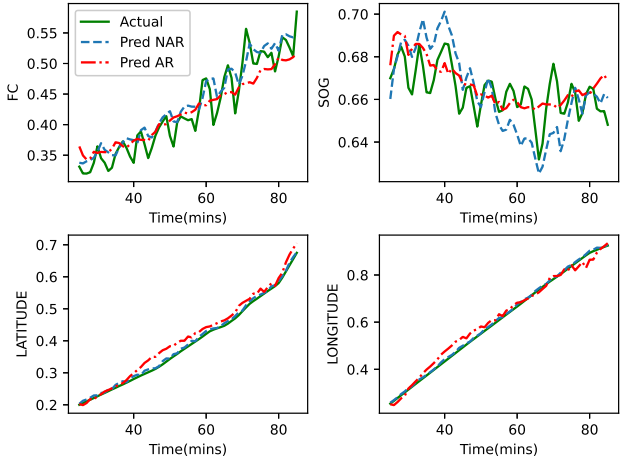

Figure 1: AR vs NAR prediction comparison

For our modeling we used PyTorch v.2.0.1. To achieve optimal model performance, we have carefully fine-tuned a selection of hyperparameters. During training, we utilized a batch size of 256 and ran for a total of 50 epochs. We initiated training with the Adam optimizer, setting the learning rate at 1e-4, and implemented a learning rate decay strategy, halving the learning rate every 5 epochs to facilitate convergence and enhance the model's adaptability.

As illustrated in Fig. 1, we conduct a comparative analysis of auto-regressive (AR) and non-auto-regressive (NAR) approaches for a random trip, specifically focusing on the first prediction horizon. In the AR approach, we employ the last predicted values as the previous states for subsequent time steps, replicating the scenario in the offline simulator, which may lead to cumulative errors. In contrast, the NAR approach substitutes the actual value as the previous state at each time step, emulating the situation in the real-time online simulator on-site. Both approaches serve as low-pass filters to mitigate high-frequency measurement noise. Remarkably, even in the case of AR, where the Gym environment lacks access to actual measured data, it still exhibits relatively strong performance compared to the measured data.

## Environment

In this section, we have established a generic reinforcement learning (RL) dataset and constructed a comprehensive Gym environment for experimentation and evaluation. During specific trips, the vessel encountered adverse sea conditions, necessitating immediate actions to avert potential hazards. In response, the captain transitioned to operation mode 2, implemented rerouting, and reduced speed as precautionary measures to ensure safe navigation. Recognizing the distinct nature of these trips compared to routine operations, we labeled them as adversarial cases. This labeling enriches the dataset, offering a more comprehensive and realistic representation of the maritime environment. However, for the scope of our current work, we exclusively focus on normal trips. This deliberate choice allows us to establish a foundational understanding and framework for our predictive and optimization models. Subsequently, we plan to expand our research to encompass adversarial scenarios, enhancing the refinement and validation of our models within a context that closely mimics real-world challenges and responses.

### RL Dataset

The RL dataset is structured as a list of trip dictionaries, with keys corresponding to observations, next observations, actions, rewards, and termination. Observations encompass a set of states for each trip, represented as matrices with dimensions $[N, Ns]$. Here, $N$ represents the total sample size within each trip, while $Ns$ signifies the number of features included. Next observations, denoted as $[N, Ns]$, encompass the subsequent observations within the trips. Actions, with dimensions $[N, Na]$, where $Na$ incorporate the size of the action space, encompassing engine speed, heading, and operation mode. Notably, we treat operation mode as a distinct action in this framework, as it exhibits an independent nature and exerts significant influence on shaping the vessel's operational characteristics. Thus, operation mode becomes an essential and dynamic component of our action space, integral to our modeling and optimization objectives. Rewards, represented as $[N, 3]$, encompass three types of rewards evaluating the actions. The first reward calculates the Euclidean distance of LAT and LON concerning the average top trip at each time step, offering a reward if the vessel is sufficiently close to the top 1% trips. The second reward solely considers FC, assigning a negative reward for each step. The third reward incorporates two components: a timeout penalty and a final reward of 1 at the destination point. The termination variable, represented as $[N, 1]$, is defined such that for all elements except the last one its value is set to False.

$$\begin{aligned}
\mathcal{D} &\equiv \sqrt{(LAT - LAT^*)^2 + (LON - LON^*)^2} \\
R_1 &= \begin{cases} -\mathcal{D} & \text{if } \mathcal{D} < 0.05 \\ 0 & \text{o.w.} \end{cases} \\
R_2 &= -FC \\
R_3 &= \begin{cases} -0.1 * \lfloor (t-90)/10 \rfloor & \text{if } t > 100 \\ 1 & \text{if } D^* < 0.05 \end{cases}
\end{aligned} \quad (3)$$

where $LAT^*$ and $LON^*$ represent the averages of the top 1% of the dataset in terms of fuel efficiency, $D^*$ denotes the Euclidean distance to the corresponding docking point, $t$ represents the time within the trip, and $\lfloor . \rfloor$ is integer floor operation. The $R_1, R_2, R_3$ are the different rewards.

### Gym environment

Building upon our offline dataset and the trained model, we have developed a Gym environment as show in Fig. 2, capable of initiating trips and disturbance factors based on real data, taking various actions to predict states in subsequent time steps, and evaluating these actions by calculating corresponding rewards. This Gym environment serves as a valuable platform for our future optimization efforts, allowing us to conduct evaluations and test strategies and algorithms within a controlled yet realistic maritime simulator.

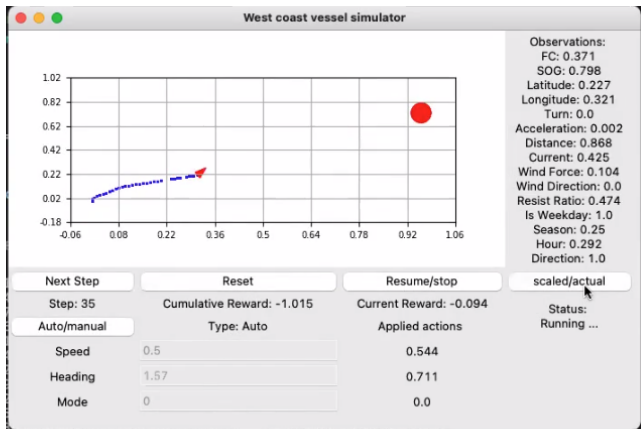

Figure 2: Simulator V0.1

Within this Gym environment, we have designed a set of methods to facilitate interactions with the simulator. The primary methods for interaction include Reset, Step, and Render. The Reset method resets the environment to its initial state, preparing it for interaction by initializing observations for a new trip drawn from the dataset. The Step method takes an action set (heading, speed, mode) as input and returns next observations, rewards, a "done" flag, and a termination flag. It predicts the next values of FC, SOG, LAT, and LON using our trained model, generates the next observation based on the input actions and predicted values, calculates the reward for the current step based on a predefined reward structure as described earlier, and assesses whether the vessel has reached its designated docking area or if a timeout condition has been met, signifying the end of the current trip. The Render method serves as the visual component of our Gym environment, enabling users to interact with and visualize simulated maritime scenarios. It provides a graphical representation of the current location and trajectory of the vessel, enhancing users' understanding of the ongoing simulation.

## Discussion

In the proposed architecture, the GRU unit will produce a correction on the output of the informer. During each iteration, the preceding iteration's output of GRU is concatenated with the informer output following the application of fully connected layers. This concatenated output serves the dual purpose of predicting an update on the informer output while simultaneously assuming the role of the hidden layer for the subsequent iteration. This iterative process enables a gradual improvement in predictions, enhancing their accuracy and reliability across multiple continuous time series forecasts. Importantly, this refinement is achieved without the need for prior knowledge of the actual values.



\end{document}